\pdfoutput=1

\documentclass[11pt]{article}

\usepackage{acl}

\usepackage{lmodern}
\usepackage{anyfontsize}

\usepackage{times}
\usepackage{latexsym}

\usepackage[T1]{fontenc}

\usepackage[utf8]{inputenc}

\usepackage{microtype}

\usepackage{algorithm,algorithmic,letltxmacro}
\usepackage{amsmath}
\usepackage{amsfonts}
\usepackage{booktabs}
\usepackage{multirow}
\usepackage{graphicx}
\usepackage{CJKutf8}
\usepackage{color}
\usepackage{tikz}
\usepackage{amssymb}
\usepackage{subcaption}
\usepackage{mathpartir}
\usepackage{forest}
\forestset{
dg edges/.style={for tree={parent anchor=south, child anchor=north,align=center,base=bottom,where n children=0{tier=word,edge=dotted,calign with current edge}{}}},
}

\usepackage{tikz-dependency}
\usepackage{mathtools}
\usepackage{hyperref}
\usepackage{mathabx}
\usepackage{tablefootnote}
\usepackage{colortbl}
\usepackage{arydshln}
\usepackage{stmaryrd}
\usepackage{wasysym}
\usepackage{fontawesome}
\usepackage{pgfplots}
\pgfplotsset{compat=1.18}
\usetikzlibrary[angles, arrows.meta, decorations.pathmorphing, decorations.pathreplacing, backgrounds, intersections, positioning, fit, petri, graphs, math, external, calc, shapes, shapes.geometric, decorations.text, shadings, pgfplots.groupplots]
\usepackage{cleveref}

\definecolor{brickred}{HTML}{b92622}
\definecolor{midnightblue}{HTML}{005c7f}
\definecolor{salmon}{HTML}{f1958d}
\definecolor{burntorange}{HTML}{3B2928} 
\definecolor{junglegreen}{HTML}{4dae9d}
\definecolor{forestgreen}{HTML}{499c5e}
\definecolor{pinegreen}{HTML}{3d8a75}
\definecolor{seagreen}{HTML}{6bc1a2}
\definecolor{limegreen}{HTML}{97c65a}

\newcommand{\brickred}[1]{\textcolor{brickred}{#1}}

\newcommand{\salmon}[1]{\textcolor{salmon}{#1}}

\DeclareDocumentCommand{\trapezoid}{O{2.0} O{1.0} O{0.5} m m}{
    \begin{scope}[scale=0.9,thick]
        \draw[anchor=mid] (0, 0) -- (0, -{#2}) node[below=9pt,anchor=base] {\ensuremath{#4}} -- ({#1}, -{#2}) node [below=9pt,anchor=base] {\ensuremath{#5}} -- ({#1}, -{#3}) -- cycle;
    \end{scope}
}

\DeclareDocumentCommand{\trapezoidd}{O{2.0} O{1.0} O{0.5} m m m}{
    \begin{scope}[scale=0.9,thick]
        \draw[anchor=mid] (0, 0) -- (0, -{#2}) node[below=9pt,anchor=base] {\ensuremath{#4}} -- ({#1}, -{#2}) node [below=9pt,anchor=base] {\ensuremath{#5}} -- ({#1}, -{#3}) -- cycle;
            \draw node[] at ( $({#2}, 0.2)!0.5!(0, 0.2)$) {\ensuremath{#6}};
    \end{scope}
}

\DeclareDocumentCommand{\trapezoiddd}{O{2.0} O{1.0} O{0.5} m m m m}{
    \begin{scope}[scale=0.9,thick]
        \draw[anchor=mid] (0, 0) -- (0, -{#2}) node[below=9pt,anchor=base] {\ensuremath{#4}} -- ({#1}, -{#2}) node [below=9pt,anchor=base] {\ensuremath{#5}} -- ({#1}, -{#3}) -- cycle;
            \draw node[] at ( $({#2}, 0.7)!0.5!(0, 0.7)$) {\ensuremath{#6}};
            \draw node[] at ( $({#2}, 0.2)!0.5!(0, 0.2)$) {\ensuremath{#7}};
    \end{scope}
}

\DeclareDocumentCommand{\trapezoiddleft}{O{2.0} O{1.0} O{0.5} m m m}{
    \begin{scope}[scale=0.9,thick]
        \draw[anchor=mid] (0, 0) -- (0, -{#2}) node[below=9pt,anchor=base] {\ensuremath{#4}} -- ({#1}, -{#2}) node [below=9pt,anchor=base] {\ensuremath{#5}} -- ({#1}, -{#3}) -- cycle;
            \node at ( $({#2}, 0.5)!0.5!(0, 0.5)$) [] {\ensuremath{#6}};
    \end{scope}
}

\DeclareDocumentCommand{\trapezoiddleftt}{O{2.0} O{1.0} O{0.5} m m m m}{
    \begin{scope}[scale=0.9,thick]
        \draw[anchor=mid] (0, 0) -- (0, -{#2}) node[below=9pt,anchor=base] {\ensuremath{#4}} -- ({#1}, -{#2}) node [below=9pt,anchor=base] {\ensuremath{#5}} -- ({#1}, -{#3}) -- cycle;
            \node at ( $({#2}, 1)!0.5!(0, 1)$) [] {\ensuremath{#6}};
                        \node at ( $({#2}, 0.5)!0.5!(0, 0.5)$) [] {\ensuremath{#7}};
    \end{scope}
}

\DeclareDocumentCommand{\righttriangle}{O{0.5} O{0.5} m m }{
    \begin{scope}[scale=0.9,thick]
        \draw[anchor=mid] (0, 0) -- (0, -{#2}) node[below=9pt,anchor=base] {\ensuremath{#3}} -- ({#1}, -{#2}) node [below right=9pt and 3pt,anchor=base] {\ensuremath{#4}} -- cycle;
    \end{scope}
}

\DeclareDocumentCommand{\righttrianglee}{O{0.5} O{0.5} m m m}{
    \begin{scope}[scale=0.9,thick]
        \draw[anchor=mid](0, 0) -- (0, -{#2}) node[below=9pt,anchor=base] {\ensuremath{#3}} -- ({#1}, -{#2}) node [below right=9pt and 3pt,anchor=base] {\ensuremath{#4}} -- cycle;
        \draw node[] at ( $({#1}, 0.2)!0.5!(0, 0.2)$) {\ensuremath{#5}};
    \end{scope}
}

\DeclareDocumentCommand{\lefttriangle}{O{0.5} O{0.5} m m}{
    \begin{scope}[scale=0.9,thick]
        \draw[anchor=mid] (0, -{#2}) node[below left=9pt and 3pt,anchor=base] {\ensuremath{#3}} -- ({#1}, -{#2}) node [below=9pt,anchor=base] {\ensuremath{#4}} -- ({#1}, 0) -- cycle;
    \end{scope}
}

\DeclareDocumentCommand{\square}{O{0.5} O{0.5} m m m}{
    \begin{scope}[scale=0.9,thick]
        \draw[anchor=mid] (0, -{#2}) node[below left=9pt and 3pt,anchor=base] {\ensuremath{#3}} -- ({#1}, -{#2}) node [below=9pt,anchor=base] {\ensuremath{#4}} -- ({#1}, 0) --  (0, 0) -- cycle;
                \draw node[] at ( $({#1}, 0.2)!0.5!(0, 0.2)$) {\ensuremath{#5}};
    \end{scope}
}

\DeclareDocumentCommand{\leftclosetriangle}{O{0.5} O{0.5} m m}{
    \begin{scope}[scale=0.9,thick]
        \draw[anchor=mid] (0, -{#2}) node[below left=9pt and 3pt,anchor=base] {\ensuremath{#3}} -- ({#1}, -{#2}) node [below=9pt,anchor=base] {\ensuremath{#4}} -- ({#1}, 0) -- cycle;
        \draw[anchor=mid] ({#1}, -{#2}-0.05) -- (0, -{#2}-0.05);
    \end{scope}
}

\DeclareDocumentCommand{\leftclosetrianglee}{O{0.5} O{0.5} m m m}{
    \begin{scope}[scale=0.9,thick]
        \draw[anchor=mid] (0, -{#2}) node[below left=9pt and 3pt,anchor=base] {\ensuremath{#3}} -- ({#1}, -{#2}) node [below=9pt,anchor=base] {\ensuremath{#4}} -- ({#1}, 0) -- cycle;
        \draw[anchor=mid] ({#1}, -{#2}-0.05) -- (0, -{#2}-0.05);
        \draw node[] at ( $({#1}, 0.2)!0.5!(0, 0.2)$) {\ensuremath{#5}};
    \end{scope}
}

\DeclareDocumentCommand{\lefttrianglee}{O{0.5} O{0.5} m m m}{
    \begin{scope}[scale=0.9,thick]
        \draw[anchor=mid] (0, -{#2}) node[below left=9pt and 3pt,anchor=base] {\ensuremath{#3}} -- ({#1}, -{#2}) node [below=9pt,anchor=base] {\ensuremath{#4}} -- ({#1}, 0) -- cycle;
        \draw node[] at ( $({#1}, 0.2)!0.5!(0, 0.2)$) {\ensuremath{#5}};
    \end{scope}
}

\DeclareDocumentCommand{\rightclosetriangle}{O{0.5} O{0.5} m m}{
    \begin{scope}[scale=0.9,thick]
        \draw[anchor=mid] (0, 0) -- (0, -{#2}) node[below=9pt,anchor=base] {\ensuremath{#3}} -- ({#1}, -{#2}) node [below right=9pt and 3pt,anchor=base] {\ensuremath{#4}} -- cycle;
        \draw[anchor=mid] ({0}, -{#2}-0.05) -- ({#1}, -{#2}-0.05);

    \end{scope}
}

\DeclareDocumentCommand{\rightclosetrianglee}{O{0.5} O{0.5} m m m}{
    \begin{scope}[scale=0.9,thick]
        \draw[anchor=mid] (0, 0) -- (0, -{#2}) node[below=9pt,anchor=base] {\ensuremath{#3}} -- ({#1}, -{#2}) node [below right=9pt and 3pt,anchor=base] {\ensuremath{#4}} -- cycle;
        \draw[anchor=mid] ({0}, -{#2}-0.05) -- ({#1}, -{#2}-0.05);
        \draw node[] at ( $({#1}, 0.2)!0.5!(0, 0.2)$) {\ensuremath{#5}};
    \end{scope}
}

\DeclareDocumentCommand{\triangle}{O{0.5} O{0.5} m m m m}{
    \begin{scope}[scale=0.9,thick]
        \draw[anchor=mid] (0, 0) node[draw, circle, black, fill, scale=0.6]{} -- (-{#1}, -{#2}) node[below=9pt,anchor=base] {\ensuremath{#3}}  --  (0, -{#2}) node[below=9pt,anchor=base] {\ensuremath{#4}} -- ({#1}, -{#2}) node [below right=9pt and 3pt,anchor=base] {\ensuremath{#5}} -- cycle;
        \draw node at ( 0, 0.5) {\ensuremath{#6}};
    \end{scope}
}

\DeclareDocumentCommand{\triangleclose}{O{0.5} O{0.5} m m m m}{
    \begin{scope}[scale=0.9,thick]
        \draw[anchor=mid] (0, 0) node[draw, circle, black, fill, scale=0.6]{} -- (-{#1}, -{#2}) node[below=9pt,anchor=base] {\ensuremath{#3}}  --  (0, -{#2}) node[below=9pt,anchor=base] {\ensuremath{#4}} -- ({#1}, -{#2}) node [below right=9pt and 3pt,anchor=base] {\ensuremath{#5}} -- cycle;
        \draw[anchor=mid] (-{#1}, -{#2}-0.05) -- ({#1}, -{#2}-0.05);
        \draw node[] at ( 0, 0.5) {\ensuremath{#6}};
    \end{scope}
}

\DeclareDocumentCommand{\triangleleft}{O{0.5} O{0.5} O{0.5} m m m m}{
    \begin{scope}[scale=0.9,thick]
        \draw[anchor=mid] (0, 0) node[draw, circle, black, fill, scale=0.6](head){} -- (-{#1}, -{#2}) node[below=9pt,anchor=base] {\ensuremath{#4}}  --  (0, -{#2}) node[below=9pt,anchor=base] {\ensuremath{#5}} -- ({#1}, -{#2}) node [below right=9pt and 3pt,anchor=base] {\ensuremath{#6}} -- cycle;
        \draw node[draw, circle, black, fill, scale=0.4] at (-{#3}, -{#2}) (node1) {};
        \draw node[below=9pt, anchor=base] at (-{#3}, -{#2}) {$h$};
                \draw[anchor=mid] (-{#1}, -{#2}-0.05) -- ({#1}, -{#2}-0.05);
        \draw[->] (node1) to[out=90] (head);
        \draw node[] at ( 0, 0.5) {\ensuremath{#7}};
        
    \end{scope}
}

\DeclareDocumentCommand{\triangleright}{O{0.5} O{0.5} O{0.5} m m m m}{
    \begin{scope}[scale=0.9,thick]
        \draw[anchor=mid] (0, 0) node[draw, circle, black, fill, scale=0.6](head){} -- (-{#1}, -{#2}) node[below=9pt,anchor=base] {\ensuremath{#4}}  --  (0, -{#2}) node[below=9pt,anchor=base] {\ensuremath{#5}} -- ({#1}, -{#2}) node [below right=9pt and 3pt,anchor=base] {\ensuremath{#6}} -- cycle;
        \draw node[draw, circle, black, fill, scale=0.4] at ({#3}, -{#2}) (node1) {};
        \draw node[below=9pt, anchor=base] at ({#3}, -{#2}) {$h$};
                \draw[anchor=mid] (-{#1}, -{#2}-0.05) -- ({#1}, -{#2}-0.05);
        \draw[->] (node1) to[in=45,out=90] (0,0);
        \draw node[] at ( 0, 0.5) {\ensuremath{#7}};
        
    \end{scope}
}

\title{Revisiting Structured Sentiment Analysis\\ as Latent Dependency Graph Parsing}

\author{Chengjie Zhou\textsuperscript{\rm 1},  \, Bobo Li\textsuperscript{\rm 1},  \,Hao Fei\textsuperscript{\rm 2},   \,Fei Li\textsuperscript{\rm 1},  \,Chong Teng\textsuperscript{\rm 1},   \,Donghong Ji\textsuperscript{\rm 1}\Thanks{ Corresponding author} \\
\textsuperscript{\rm 1} Key Laboratory of Aerospace Information Security and Trusted Computing, Ministry of \\
Education, School of Cyber Science and Engineering, Wuhan University, Wuhan, China \\
\textsuperscript{\rm 2} National University of Singapore, Singapore \\
\texttt{\{zhoucj,boboli,lifei\_csnlp,tengchong,dhji\}@whu.edu.cn}}

\begin{document}
\maketitle
\renewcommand{\thefootnote}{\fnsymbol{footnote}}
\renewcommand{\thefootnote}{\arabic{footnote}}
\begin{abstract}
Structured Sentiment Analysis (SSA) was cast as a problem of bi-lexical dependency graph parsing by prior studies.
Multiple formulations have been proposed to construct the graph, which share several intrinsic drawbacks:
(1) The internal structures of spans are neglected, thus only the boundary tokens of spans are used for relation prediction and span recognition, thus hindering the model's expressiveness;
(2) Long spans occupy a significant proportion in the SSA datasets, which further exacerbates the problem of internal structure neglect.
In this paper, we treat the SSA task as a dependency parsing task on partially-observed dependency trees, regarding flat spans without determined tree annotations as latent subtrees to consider internal structures of spans.
We propose a two-stage parsing method and leverage TreeCRFs with a novel constrained inside algorithm to model latent structures explicitly, which also takes advantages of joint scoring graph arcs and headed spans for global optimization and inference. 
Results of extensive experiments on five benchmark datasets reveal that our method performs significantly better than all previous bi-lexical methods, achieving new state-of-the-art.

\end{abstract}

\section{Introduction}\label{sec:intro}

\begin{figure*}[ht!]
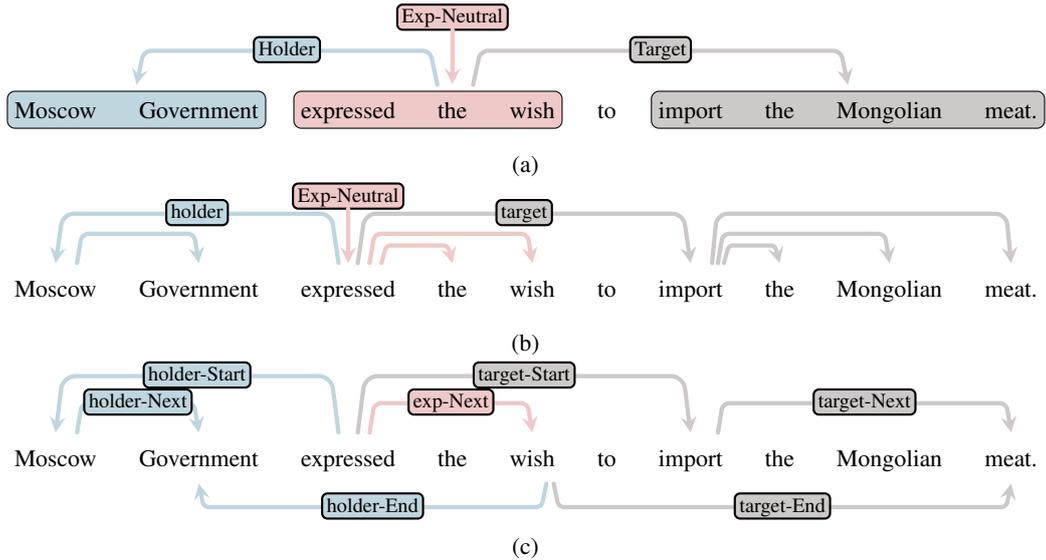

    \centering
    
    \begin{subfigure}{\textwidth}
    \centering
    
    \begin{dependency}[scale=0.9]
        \begin{deptext}[column sep=1em,font=\small]
        Moscow\& Government \& expressed \&the \&wish \&to \&import \&the \&Mongolian \&meat. \\
        \end{deptext}

        \wordgroup[fill=midnightblue!25,  rounded corners=1mm]{1}{1}{2}{holder}
        \wordgroup[fill=brickred!25, rounded corners=1mm]{1}{3}{5}{exp}
        \wordgroup[fill=burntorange!25, rounded corners=1mm]{1}{7}{10}{target}
        \groupedge[->, ultra thick, midnightblue!25, label style={fill= midnightblue!25,thick},edge start x offset=1.6ex]{exp}{holder}{ Holder}{.5cm}
        \groupedge[->, ultra thick, burntorange!25, label style={fill= burntorange!25,thick},edge start x offset=3ex]{exp}{target}{ Target}{.5cm}
        \deproot[edge vertical padding=0.6ex, edge height=7ex, label style={fill=brickred!25, thick}, edge style={brickred!25,ultra thick}]{4}{{Exp-Neutral}}
    \end{dependency}
    \caption{}
    
    \end{subfigure}
    
    \begin{subfigure}[!h]{\textwidth}
        \centering
         \begin{dependency}[scale=0.9]
            \begin{deptext}[column sep=1em,font=\small]
                 Moscow\& Government \& expressed \&the \&wish \&to \&import \&the \&Mongolian \&meat. \\
            \end{deptext}

            \deproot[edge vertical padding=0.6ex, edge height=7ex, label style={fill=brickred!25, thick}, edge style={brickred!25,ultra thick}]{3}{{Exp-Neutral}}
            \depedge[edge vertical padding=0.6ex, edge height=4.2ex, label style={fill=midnightblue!25, thick},  ultra thick,midnightblue!25]{3}{1}{holder}
            \depedge[edge vertical padding=0.6ex, edge height=4.2ex, label style={fill=burntorange!25, thick}, ultra thick,burntorange!25]{3}{7}{target}
            \depedge[->, edge vertical padding=0.6ex, edge start x offset=1ex, edge height=2.5ex, edge slant=2pt, hide label,ultra thick, brickred!25]{3}{5}{}
            \depedge[->, edge vertical padding=0.6ex, edge start x offset=2ex, edge height=1.5ex, edge slant=2pt, hide label, ultra thick, brickred!25]{3}{4}{}
            \depedge[->, edge vertical padding=0.6ex, edge start x offset=1ex, edge height=2.5ex, edge slant=2pt, hide label,ultra thick, midnightblue!25]{1}{2}{}
            \depedge[->, edge vertical padding=0.6ex, edge start x offset=2ex, edge height=1.5ex, edge slant=2pt, hide label,ultra thick, burntorange!25]{7}{8}{}
            \depedge[->, edge vertical padding=0.6ex, edge start x offset=1.5ex, edge height=2.5ex, edge slant=2pt, hide label,ultra thick, burntorange!25]{7}{9}{}
            \depedge[->, edge vertical padding=0.6ex, edge start x offset=1ex, edge height=4.2ex, edge slant=2pt, hide label,ultra thick, burntorange!25]{7}{10}{}

        \end{dependency}
        \caption{}
        
    \end{subfigure}
    \vspace{4mm}
    \begin{subfigure}[!h]{\textwidth}
        \centering
         \begin{dependency}[scale=0.9]
            \begin{deptext}[column sep=1em,font=\small]
                 Moscow\& Government \& expressed \&the \&wish \&to \&import \&the \&Mongolian \&meat. \\
            \end{deptext}

            \depedge[edge vertical padding=0.6ex, edge height=4.8ex, label style={fill=midnightblue!25, thick},  ultra thick,midnightblue!25]{3}{1}{holder-Start}
            \depedge[edge vertical padding=0.6ex, edge height=4.8ex, label style={fill=burntorange!25, thick}, ultra thick,burntorange!25]{3}{7}{target-Start}
            \depedge[->, edge vertical padding=0.6ex, label style={fill=brickred!25, thick}, edge start x offset=1ex, edge height=2.5ex, edge slant=2pt, ultra thick, brickred!25]{3}{5}{exp-Next}
            \depedge[->, edge vertical padding=0.6ex, label style={fill=midnightblue!25, thick}, edge start x offset=1ex, edge height=2.5ex, edge slant=2pt, ultra thick, midnightblue!25]{1}{2}{holder-Next}
            \depedge[->, edge below,edge vertical padding=0.6ex, label style={fill=midnightblue!25, thick}, edge start x offset=2ex, edge height=2ex, edge slant=2pt, ultra thick, midnightblue!25]{5}{2}{holder-End}
            \depedge[->, edge vertical padding=0.6ex, label style={fill=burntorange!25, thick}, edge start x offset=1.5ex, edge height=2.5ex, edge slant=2pt, ultra thick,burntorange!25]{7}{10}{target-Next}
            \depedge[->,edge below, edge vertical padding=0.6ex, label style={fill=burntorange!25, thick}, edge start x offset=1ex, edge height=2ex, edge slant=2pt,ultra thick, burntorange!25]{5}{10}{target-End}

        \end{dependency}
        \caption{}
    \end{subfigure}
    \caption{(\textbf{a}) An example of original structure sentiment analysis. 
    (\textbf{b}) The head-first parsing graph proposed by \citet{barnes_structured_2021}.
    (\textbf{c}) The label strategy proposed by \citet{shi_effective_2022}. 
    The label formulation proposed by \citet{zhai_ussa_2023} is similar to this. 
    }
    \label{fig1:example}
\end{figure*}

Structured Sentiment Analysis (SSA) aims to extract the complete opinion tuple from a sentence. 
As shown in Figure~\ref{fig1:example}(a), the complete opinion tuple includes an opinion expression \textbf{$e$} with sentiment polarity \textbf{$p$}, an opinion holder \textbf{$h$}, and the corresponding target \textbf{$t$}.
Given the complexity of detecting three items and classifying one, SSA presents more challenges than other related tasks, such as Opinion Mining \citep{katiyar_investigating_2016,xia_unified_2021}, ABSA (Aspect-based Sentiment Analysis) \citep{pontiki_semeval-2014_2014,pontiki_semeval-2016_2016,wang_recursive_2016}, TOWE (Target-oriented Opinion Words Extraction) \citep{fan_target-oriented_2019,mao_joint_2021}, ASTE (Aspect Sentiment Triplet Extraction) \citep{peng_knowing_2020,mao_joint_2021,zhai_com-mrc_2022,Feiijcai22UABSA,li-etal-2023-diaasq,li2024harnessing}, etc.

Recent works of SSA mainly cast it as a problem of bi-lexical dependency graph parsing and propose multiple formulations:
(1)~\citet{barnes_structured_2021} proposed formulations namely head-first/head-final as illustrated in Figure \ref{fig1:example}(b). 
Their method cannot resolve the problem because head-first/head-final treats the first/final word as the head of the span and strictly restricts any word inside the span directly head to span head, which decreases the height of the converted trees to 2 and excludes the latent structures completely.
(2) Another label strategy was proposed by \citet{shi_effective_2022}, 
which simplifies the label set to only arcs linking spans boundaries, as shown in Figure~\ref{fig1:example}(c). 
Despite the special label for discontinuous span decoding, ~\citet{zhai_ussa_2023} utilize the same label set.
Without distinct formulation about inside words, they attempt to utilize the powerful neural models like Graph Attention Network \citep{velickovic_graph_2018} or Axial-based Attention Network \citep{huang_ccnet_2019,wang_axial-deeplab_2020} to implicitly encoder the inside structure information, which is found lagging behind large with explicitly modeling with graph-based parsing methods \citep{wang_second-order_2020,fonseca_revisiting_2020,zhang_efficient_2020,yang_combining_2022}. 
It is evident that previous work does not address the key challenge focused on the prediction of boundary words (First/Final/Both) of spans, and neglect the words and structures inside spans, which hinders the model expressiveness seriously.

\begin{table}[!t]
\footnotesize
\renewcommand{\arraystretch}{1}
\setlength{\tabcolsep}{3mm}
\centering
\begin{tabular}{@{}lrrrc@{}}
\toprule
\multicolumn{1}{c}{
\multirow{2}{*}{Dataset}} & 
\multicolumn{3}{c}{Span Length $\geq$ 4} & 
\multicolumn{1}{c}{\multirow{2}{*}{Max} } \\ \cmidrule(l){2-4} 
\multicolumn{1}{c}{}   & \multicolumn{1}{c}{Holder} & \multicolumn{1}{c}{Target} & \multicolumn{1}{c}{Exp.}   \\ \midrule
\textbf{NoReC}$_\textsuperscript{Fine}$      & 1.1\%                     & \brickred{\textbf{19.2\%}}                     & \brickred{\textbf{56.8\%}}            & \brickred{\textbf{40}}             \\
\textbf{MultiB}$_\textsuperscript{CA}$                                                                                     & 2.6\%                     & \brickred{\textbf{18.4\%}}                     & \brickred{\textbf{21.4\%}}           & \brickred{\textbf{19}}               \\
\textbf{MultiB}$_\textsuperscript{EU}$                                                                                    & 1.1\%                     & 2.7\%                      & \brickred{\textbf{15.3\%}}                & 10           \\
\textbf{MPQA}                                                                                   & \brickred{\textbf{19.9\%}}                    & \brickred{\textbf{51.1\%}}                     & 14.5\%         & \brickred{\textbf{56}}                 \\
\textbf{DS}$_\textsuperscript{Unis}$                                                                                   & 1.3\%                     & 0.8\%                      & \brickred{\textbf{13.7\%}}         & 9                   \\ \bottomrule
\end{tabular}
\caption{Statistics of the proportion of span 1) with length greater than or equal to 4 and 2) the max length of span (both in tokens) for SSA datasets. We highlight the number that reveals that the long span is ubiquitous and the extraction of them seems to be the bottleneck of SSA problem.}
\label{Table1:span-length}
\end{table}

Should we neglect the structure inside spans as previous works have done? 
Table \ref{Table1:span-length} list the statistics of span length and the max length of spans in the benchmark datasets respectively. 
We present a real example to illustrate our point, focusing on the expression ``conceded'' and the target ``US President ... last year'' for brevity.
\begin{quote}
    Tang \brickred{\emph{conceded}} there had been ``twists and turns'' following \salmon{\emph{US President George W. Bush 's accession to the US presidency in early last year}}.
\end{quote}
From this example, it is evident that the target span is considerably long, making it challenging for the boundaries (``US'' and ``year'') to effectively represent the entire span.
Conversely, the internal word ``accession'' provides significant clues for identifying it as the target of the expression ``conceded''. Furthermore, the other words within the target span act as modifiers, aiding in the detection of the span's boundaries.
Based on these observations, the necessity of addressing the internal structure of spans in the SSA task is clear. It remains a significant challenge to develop an effective and unified method that can handle such structures.

To address this issue, we propose a novel approach to structured sentiment analysis: treating flat spans as latent subtrees. This perspective considers expression-holder/target structures as partially-observed trees, where the exact subtrees for each span are yet to be determined. We employ TreeCRF \citep{eisner_bilexical_1997,eisner_efficient_1999} to model these partially-observed trees. This involves enumerating all possible arcs and spans in a latent tree form of a flat span and calculating their probabilities using a constrained inside algorithm. The scores for all possible arcs and spans are determined using biaffine or triaffine attention methods \citep{Dozat2017,zhang_efficient_2020}, commonly applied in Dependency Parsing. During the decoding stage, we parse the highest-scoring dependency tree and reconstruct the SSA structure from it. Span boundaries are accurately inferred from the descendant words of a head word. Additionally, using the labels from the dependency tree, we can globally predict span-span relationships. This method not only explicitly accounts for internal span structures but also maintains the end-to-end nature characteristic of previous work.

We conduct extensive experiments on five benchmarks, including \text{NoReC}$_\textsuperscript{Fine}$ \citep{ovrelid_fine-grained_2020}, \text{MultiB}$_\textsuperscript{EU}$,  \text{MultiB}$_\textsuperscript{CA}$ \citep{barnes_multibooked_2018}, \text{MPQA} \citep{wiebe_annotating_2005} and \text{DS}$_\textsuperscript{Unis}$ \citep{toprak_sentence_2010}. 
The results affirm that our model achieves new state-of-the-art in performance for SSA task.

Our contributions are summarized as follows:
\begin{itemize}
\setlength{\itemsep}{0pt}
\setlength{\parsep}{0pt}
\setlength{\parskip}{0pt}
\item We cast SSA as a novel latent trees formulation to address the neglecting of span structures in prior work. Concretely, we treat flat spans as latent trees and marginalize the explicit structure via a novel constraint inside algorithm.

\item We propose an effective two-stage parsing method to well collaborate with our latent tree formulation, which employs dependency parsing with high-order scoring and global optimization, modeling sentiment structures explicitly.

\item The experimental results show that our model has achieved the SOTA performance in five datasets for structured sentiment analysis, especially in terms of long spans boundary detection and relation prediction.

\end{itemize}

\section{Related Work}\label{sec:related}

As a key topic in the Sentiment Analysis community \cite{fei-etal-2023-reasoning,Wu0RJL21}, Structured Sentiment Analysis (SSA) encompasses several sub-tasks, each targeting a specific component of the goal tuple (holder, target, expression, polarity).
SSA also involves closely with structure predictions \cite{FeiMatchStruICML22} and relevant tasks \cite{fei-etal-2023-scene,wu-etal-2023-cross2stra,wu2023next}.

\paragraph{OM}
Opinion Mining primarily aims to extract the (h, t, e) tuple. Existing OM studies generally adopt one of two approaches: 1) BIO-based approach, which views OM as a sequence labeling task \citep{katiyar_investigating_2016}; and 2) span-based approach, which jointly predicts all span pairs and their interrelations \citep{xia_unified_2021}. Additionally, \citet{zhang_end--end_2019,Wu0LZLTJ22} introduced a transition-based model for OM. 
However, these methodologies neglect the sentiment polarity classification sub-task.

\paragraph{ABSA}
Aspect-Based Sentiment Analysis is another important sentiment analysis task. Various methods have been proposed to address ABSA, including: 1) Pipeline \citep{peng_knowing_2020}, which sequentially predicts spans and their relationships; 2) End-to-End \citep{chen_relation-aware_2020}, utilizing interactive information from each pair of sub-tasks; and 3) MRC  \citep{mao_joint_2021,zhai_com-mrc_2022}, which employ a machine reading comprehension framework to extract triplets. 
However, these methodologies neglect the sentiment holder extraction sub-task.

\begin{figure*}[ht!]
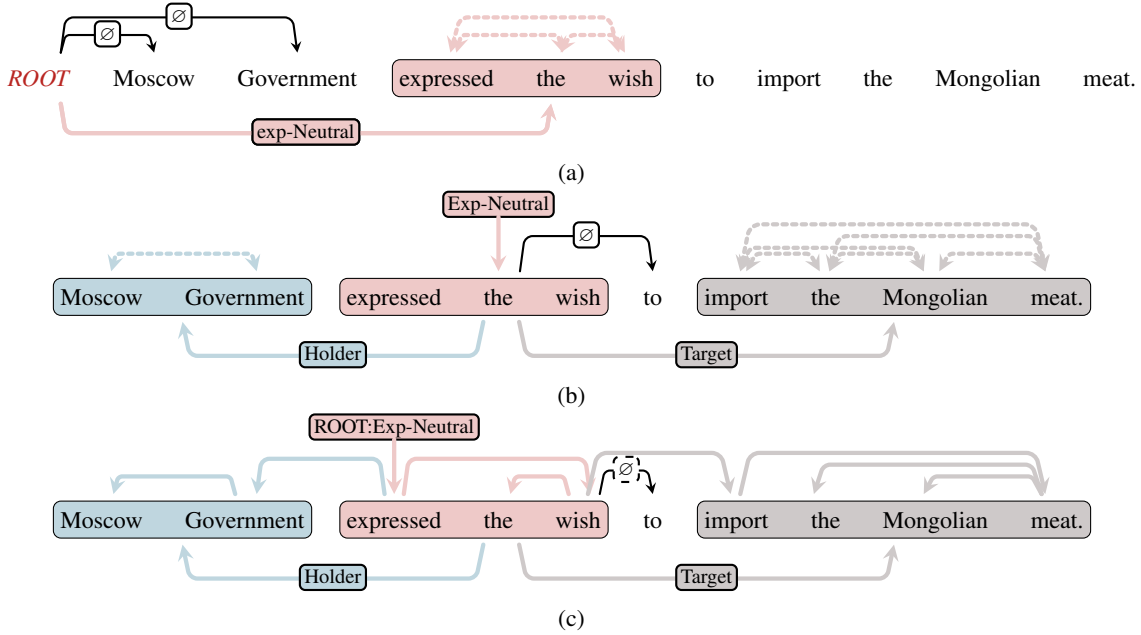

    \centering
    
    \begin{subfigure}{\textwidth}
    \centering
    
    \begin{dependency}[scale=0.9]
        \begin{deptext}[column sep=1em,font=\small]
        \brickred{\emph{ROOT}}\& Moscow\& Government \& expressed \&the \&wish \&to \&import \&the \&Mongolian \&meat. \\
        \end{deptext}
        \depedge[<->, edge vertical padding=0.6ex, edge start x offset=0ex, edge height=1.5ex, edge end x offset=1ex,edge slant=2pt, hide label, ultra thick, brickred!25, dotted]{4}{5}{}
        \depedge[<->, edge vertical padding=0.6ex, edge start x offset=0ex, edge end x offset=-0.5ex, edge height=1.5ex, edge slant=4pt, hide label, ultra thick, brickred!25, dotted]{5}{6}{}
        \depedge[<->, edge vertical padding=0.6ex, edge start x offset=-0.5ex, edge end x offset=-1ex, edge height=3ex, edge slant=4pt, hide label,ultra thick, brickred!25, dotted]{4}{6}{}
        
        \wordgroup[fill=brickred!25, rounded corners=1mm]{1}{4}{6}{exp}
        \depedge[->,edge below, edge vertical padding=0.6ex, edge start x offset=1ex, edge height=2.5ex, edge slant=2pt,ultra thick, brickred!25,label style={fill=brickred!25, thick}]{1}{5}{exp-Neutral}
        \depedge[->,edge below, edge vertical padding=0.6ex, edge start x offset=1ex, edge height=2.5ex, edge slant=2pt,ultra thick, brickred!25,label style={fill=brickred!25, thick}]{1}{5}{exp-Neutral}
        \depedge[->, edge vertical padding=0.6ex, edge start x offset=1ex, edge height=1.5ex, edge slant=2pt, thick,label style={ thick}]{1}{2}{$\varnothing$}
        \depedge[->, edge vertical padding=0.6ex, edge start x offset=1ex, edge height=3ex, edge slant=2pt,thick,label style={ thick}]{1}{3}{$\varnothing$}
    \end{dependency}
    \caption{}
    
    \end{subfigure}
    
    \begin{subfigure}[!h]{\textwidth}
        \centering
         \begin{dependency}[scale=0.9]
            \begin{deptext}[column sep=1em,font=\small]
                 Moscow\& Government \& expressed \&the \&wish \&to \&import \&the \&Mongolian \&meat. \\
            \end{deptext}
            \wordgroup[fill=midnightblue!25,  rounded corners=1mm]{1}{1}{2}{holder}
            \wordgroup[fill=brickred!25, rounded corners=1mm]{1}{3}{5}{exp}
            \wordgroup[fill=burntorange!25, rounded corners=1mm]{1}{7}{10}{target}
            \groupedge[->, edge below,ultra thick, midnightblue!25, label style={fill= midnightblue!25,thick},edge start x offset=1.6ex]{exp}{holder}{ Holder}{.5cm}
            \groupedge[->, edge below,ultra thick, burntorange!25, label style={fill= burntorange!25,thick},edge start x offset=3ex]{exp}{target}{ Target}{.5cm}
            \deproot[edge vertical padding=0.6ex, edge height=7ex, label style={fill=brickred!25, thick}, edge style={brickred!25,ultra thick}]{4}{{Exp-Neutral}}
            \depedge[->, edge vertical padding=0.6ex, edge start x offset=1ex, edge height=3ex, edge slant=2pt,thick,label style={ thick}]{4}{6}{$\varnothing$}
            \depedge[<->, edge vertical padding=0.6ex, edge start x offset=0ex, edge height=1.5ex, edge end x offset=1ex,edge slant=2pt, hide label, ultra thick, midnightblue!25, dotted]{1}{2}{}
            \depedge[<->, edge vertical padding=0.6ex, edge start x offset=0ex, edge end x offset=-0.5ex, edge height=1.5ex, edge slant=4pt, hide label, ultra thick, burntorange!25, dotted]{7}{8}{}
            \depedge[<->, edge vertical padding=0.6ex, edge start x offset=-0.5ex, edge end x offset=-1ex, edge height=1.5ex, edge slant=4pt, hide label,ultra thick, burntorange!25, dotted]{8}{9}{}
            \depedge[<->, edge vertical padding=0.6ex, edge start x offset=-0.5ex, edge end x offset=-1ex, edge height=1.5ex, edge slant=4pt, hide label,ultra thick, burntorange!25, dotted]{9}{10}{}
            \depedge[<->, edge vertical padding=0.6ex, edge start x offset=-0.5ex, edge end x offset=-1ex, edge height=3ex, edge slant=4pt, hide label,ultra thick, burntorange!25, dotted]{8}{10}{}
            \depedge[<->, edge vertical padding=0.6ex, edge start x offset=-0.5ex, edge end x offset=-1ex, edge height=2ex, edge slant=4pt, hide label,ultra thick, burntorange!25, dotted]{7}{9}{}
            \depedge[<->, edge vertical padding=0.6ex, edge start x offset=-0.5ex, edge end x offset=-1ex, edge height=4ex, edge slant=4pt, hide label,ultra thick, burntorange!25, dotted]{7}{10}{}

        \end{dependency}
        \caption{}
        \label{fig:enter-label}
    \end{subfigure}
    \vspace{4mm}
    \begin{subfigure}[!h]{\textwidth}
        \centering
         \begin{dependency}[scale=0.9]
            \begin{deptext}[column sep=1em,font=\small]
                 Moscow\& Government \& expressed \&the \&wish \&to \&import \&the \&Mongolian \&meat. \\
            \end{deptext}

            \wordgroup[fill=midnightblue!25,  rounded corners=1mm]{1}{1}{2}{holder}
            \wordgroup[fill=brickred!25, rounded corners=1mm]{1}{3}{5}{exp}
            \wordgroup[fill=burntorange!25, rounded corners=1mm]{1}{7}{10}{target}
            \groupedge[->, edge below,ultra thick, midnightblue!25, label style={fill= midnightblue!25,thick},edge start x offset=1.6ex]{exp}{holder}{ Holder}{.5cm}
            \groupedge[->, edge below,ultra thick, burntorange!25, label style={fill= burntorange!25,thick},edge start x offset=3ex]{exp}{target}{ Target}{.5cm}
            \deproot[edge vertical padding=0.6ex, edge height=7ex, label style={fill=brickred!25, thick}, edge style={brickred!25,ultra thick}]{3}{{ROOT:Exp-Neutral}}
            \depedge[->, edge vertical padding=0.6ex, edge start x offset=1ex, edge height=2ex, edge slant=2pt,thick,label style={ thick,dashed}]{5}{6}{$\varnothing$}
            \depedge[->, edge vertical padding=0.6ex, edge start x offset=0ex, edge height=3ex, edge end x offset=1ex,edge slant=2pt, hide label, ultra thick, brickred!25]{3}{5}{}
            \depedge[->, edge vertical padding=0.6ex, edge start x offset=0ex, edge height=1.5ex, edge end x offset=1ex,edge slant=2pt, hide label, ultra thick, brickred!25]{5}{4}{}
            \depedge[->, edge vertical padding=0.6ex, edge start x offset=0ex, edge height=1.5ex, edge end x offset=1ex,edge slant=2pt, hide label, ultra thick, midnightblue!25]{2}{1}{}
            \depedge[->, edge vertical padding=0.6ex, edge start x offset=0ex, edge height=3ex, edge end x offset=1ex,edge slant=2pt, hide label, ultra thick, midnightblue!25]{3}{2}{}
            \depedge[->, edge vertical padding=0.6ex, edge start x offset=0ex, edge end x offset=-0.5ex, edge height=3.5ex, edge slant=4pt, hide label, ultra thick, burntorange!25]{5}{7}{}
            \depedge[->, edge vertical padding=0.6ex, edge start x offset=-0.5ex, edge end x offset=-1ex, edge height=3.5ex, edge slant=4pt, hide label,ultra thick, burntorange!25]{7}{10}{}
            \depedge[->, edge vertical padding=0.6ex, edge start x offset=-0.5ex, edge end x offset=-1ex, edge height=1.5ex, edge slant=4pt, hide label,ultra thick, burntorange!25]{10}{9}{}
            \depedge[->, edge vertical padding=0.6ex, edge start x offset=-0.5ex, edge end x offset=-1ex, edge height=2.5ex, edge slant=4pt, hide label,ultra thick, burntorange!25]{10}{8}{}      

        \end{dependency}
        \caption{}
    \end{subfigure}
    \caption{Illustration of our SSA as Latent Dependency Graph and Two-stage Parsing.
    (\textbf{a}) The conversion and training of Stage \uppercase\expandafter{\romannumeral1}-expression extraction: convert the ROOT-expression structure to a dependency tree with (dotted) latent annotations; non-expression spans are assigned a label of $\varnothing$ for clarity. We omit the $\varnothing$ edge from ROOT to ``to import ...'' for brevity.
    (\textbf{b}) The conversion and training of corresponding Stage-\uppercase\expandafter{\romannumeral2}-holder/target extraction: with the given expression, convert the expression-holder/target structure to a dependency tree with (dotted) latent annotations and irrelevant spans assigned with $\varnothing$ as well.
    (\textbf{c}) The decoding and recovery of the converted latent dependency tree: Parsing the best dependency tree (above the sentence) and recovery the tree to SSA structure (below). We combine the separate two-stage decoding and recovery in one figure for brevity.}
    \label{fig2:example}
\end{figure*}

\paragraph{SSA}
\citet{barnes_structured_2021} 
propose the conversion named head-first/head-final and applied first-order parsing method. 
\citet{shi_effective_2022} proposed a new label strategy and apply Graph Attention Network for aggregation on span boundaries for decoding,
\citet{samuel_direct_2022} apply Transformer to predict the graph directly from the text, 
\citet{zhai_ussa_2023} add new labels to model the boundaries of discontinuous spans and apply axial-attention encoder and table filling scheme to decode the relations. 
However, all these works neglect the internal structure of the spans and rely on powerful encoders to implicitly incorporate internal span structure information. 

Different from previous work, we are the first to cast SSA task as partially-observed dependency tree and apply dependency parsing method to explicitly model the internal structures of spans.
Owing to the structural similarities, leveraging NLP tasks such as parsing proves to be an effective strategy for structured prediction. 
Our research builds upon the successes observed in partially-observed tree reduction and parsing methods across various NLP tasks
including named entity recognition (NER) \citep{yu_named_2020}, nested NER \citep{fu_nested_2021,lou_nested_2022,Li00WZTJL22,FeiLasuieNIPS22} and semantic role labeling \citep{zhang_semantic_2022}.

\section{SSA as Latent Graph Parsing Scheme}\label{sec:overview}

In the formulation of SSA as Latent Dependency Graph, we treat each sentiment span as latent tree.
To build the Dependency Graph, we deal with each expression span \textbf{separately} and assumes each of them corresponds to a single-root tree.
In this tree, the sentiment head word of the expression span serves as the dependency root, with each subtree of the corresponding holder/target span attaching to it. 
Consequently, our proposed latent dependency parsing task can be divided into two subtasks: (1) expression extraction; (2) corresponding holder/target extraction. 
Note that both of these subtasks are solved by a consistent graph-based and headed-span-based parsing method, trained jointly and decoded step-by-step, which is named as \textbf{Two-stage} Parsing
\footnote{The rationale behind Two-stage parsing is grounded in the relative independence of these subtasks:
Different from the relation prediction between spans and the boundary identification of span, 
the inside structure of expression span  is minimally influenced by the inside structure of corresponding holder/target spans and vice versa.}.

For the rest of this section, we show how our method :
(1) converting SSA to latent dependency graph;
(2) training to get span/tree structural representation; 
(3) decoding to find the best tree;
(4) recovering resulting tree to sentiment tuple. 

\subsection{Conversion and Training on Latent Tree }

Formally, given an input sentence $\boldsymbol{x}=x_1,\dots, x_n$, our object is to obtain the corresponding tree structures for each expression $e \in  \mathcal{E}$ and compose them to construct the Latent Dependency Graph ultimately.

A directed dependency tree $\boldsymbol{t}$ is defined by assigning a head $h\in\{x_0,x_1,\dots,x_n\}$, accompanied by a relation label $l\in \mathcal{
L}$ to each modifier $m \in \boldsymbol{x}$. 
Here, $x_0$ is typically positioned before $\boldsymbol{x}$, serving as the root node. 

For an expression $e \in  \mathcal{E}$ within a consecutive word span $x_i,\dots,x_j$ and assigned a sentiment label $l\in \mathcal{L}$, we constraint all potential subtrees within this span to be single-rooted at a potential headword $h$, which is not realized yet. 
This concept is illustrated in Figure~\ref{fig2:example}(a), where the sentiment label $l$ is allocated as the label of the dependency (i.e., the expression label with polarity) originating from $x_0$ to the headword. A parallel approach is adopted for non-expression spans, with the distinction of setting the label to $\varnothing$ and omitting the single-root constraint. Subsequently, we designate all corresponding latent span subtrees as descendants of $T_e$.

For a corresponding holder/target span with a consecutive word span $x_i,\dots,x_j$ and a sentiment label $l\in \mathcal{L}$, we impose a similar single-rooted constraint as with expression, as shown in Figure~\ref{fig2:example}(b).
Accordingly, the sentiment label $l$ 
denotes the label of the dependency extending from each word in $e$ to the headword. Spans deemed irrelevant are treated akin to non-expression spans.

By enumerating all possible subtrees and accumulating them together, the resultant tree sets $T_e $ and $T_{h/t}$ expand exponential in size.
To manage this during training, we develop a constrained Inside algorithm to perform the enumeration (\S~\ref{sec:inside}), 
designed to prevent the formation of illegal structures that cannot be decoded into meaningful trees.

\subsection{ Decoding and Recovery on Latent Tree}
Assuming that we have trained a parser, our next step involves recovering expression and corresponding holder/target structures after decoding/parsing the highest-scoring dependency tree.

We give illustration in Figure~\ref{fig2:example}(c), we initially identify all expression spans by obtaining the highest-scoring expression tree $\boldsymbol{t}^{\ast}$ rooted at $x_0$ through our parsing method:
\begin{equation}%
    \boldsymbol{t}^{\ast} = \arg\max_{}
    \mathrm{s}(\boldsymbol{x},\boldsymbol{t})
\end{equation}

Following this, we determine the highest-scoring corresponding tree $\boldsymbol{t}^{\ast \ast}$  for $e$ using the same algorithm:
\begin{equation}%
    \boldsymbol{t}^{\ast \ast} = \arg\max_{\substack{\boldsymbol{t}: x_0 \xrightarrow[]{p} e\in\boldsymbol{t}^{\ast}}}
    \mathrm{s}(\boldsymbol{x},\boldsymbol{t})
\end{equation}
where $\mathrm{s}(\boldsymbol{x},\boldsymbol{t})$ represents the score of the tree, with $p$ denoting the polarity label of expression spans relative to $x_0$.
The tree is constrained to have its root in one of the words in $e$. We then recover corresponding holder/target spans of the expression by transforming all subtrees headed by $e$ into flat spans. If the label $l$ of the dependency $e \rightarrow h$ is not ``$\varnothing$'' (indicating irrelevant spans), then a complete span is formed, comprising $h$ and its descendants, and is assigned $l$ as its sentiment label. The final SSA output consists of a compilation of all recovered expression spans and their respective holder/target spans.

\section{Methodology}

Following previous work on dependency parsing \citep{Dozat2017,zhang_efficient_2020,yang_combining_2022}, 
our model consists of a contextualized encoder and scoring modules.
We further propose a constraint TreeCRF to compute the probabilities of the partially-observed trees of SSA.

\subsection{Encoder}

For a given sentence $\boldsymbol{x}=x_1,x_1,\dots,x_n$, we introduce special tokens <bos> and <eos> as $x_0$ and $x_{n+1}$, respectively. The vector representation for each token $x_i\in\boldsymbol{x}$ is an amalgamation of five distinct components:
\begin{equation*}
\mathbf{e}_i = \left[\mathbf{e}^{\mathrm{word}}_i;\mathbf{e}^{\mathrm{lemma}}_i;\mathbf{e}^{\mathrm{pos}}_i;\mathbf{e}^{\mathrm{char}}_i;\mathbf{e}^{\mathrm{BERT}}_i\right]
\end{equation*}
In this composition, $\mathbf{e}_i^{\mathrm{word}}$, $\mathbf{e}_i^{\mathrm{pos}}$, and $\mathbf{e}_i^{\mathrm{lemma}}$ represent word, part-of-speech (POS), and lemma embeddings, respectively. $\mathbf{e}^{\mathrm{char}}_i$ is derived from the outputs of a CharLSTM layer \cite{lample_neural_2016}. Lastly, $\mathbf{e}^{\mathrm{BERT}}_i$ constitutes the word-level embeddings obtained through mean-pooling at the last layer of BERT \citep{bert}, specifically by averaging all subword embeddings.

Then we obtain the hidden representation and the of each vectorial token representations $x_i$ via a deep BiLSTMs \cite{gal_dropout_2016} encoder.
\begin{equation}
    \begin{aligned}
    \mathbf{h}_0,\mathbf{h}_1,\dots,\mathbf{h}_n&=\mathtt{BiLSTMs}(e_0,e_1,\dots,e_n)
    \\
\mathbf{c}_0,\mathbf{c}_1,\dots,\mathbf{c}_n&=\mathtt{BiLSTMs}(e_0,e_1,\dots,e_n)
    \end{aligned}
\end{equation}
where  $f_i$ and $b_i$ are the forward and backward hidden states of the last BiLSTM layer at position $i$ respectively, 
$h_{i} = \left[f_{i},b_{i}\right]$ is the token representation of $x_i$ ,
$c_{i} = \left[f_{i},b_{i+1}\right] $ is the boundary representation for the $i$th boundary lying between $x_i$ and $x_{i+1}$. 

\subsection{Tree Scoring}
We decompose a tree $\boldsymbol{t}$ into two distinct components: $\boldsymbol{y}$ , representing an unlabeled skeletal tree, and $\boldsymbol{l}$, signifying the corresponding sequence of labels. 
The process of scoring an unlabeled skeletal tree involves the aggregation of arcs and head-span scores.
For each head-modifier pair $h\rightarrow m \in \boldsymbol{y}$, we score them using two MLPs followed by a Biaffine layer (first-order scorer on arcs in the tree):
\begin{equation}
    \begin{aligned}
        \mathbf{r}^{\mathrm{head}/\mathrm{mod}}_i & = \mathtt{MLP}^{\mathrm{head}/\mathrm{mod}}(\mathbf{h}_i)                                \\
        \mathbf{s}_{h\rightarrow m}^{arc}                & = 
        \left[\mathbf{r}^{\mathrm{head}}_{h};1\right]^{T}W\left[\mathbf{r}^{\mathrm{mod}}_{m};1\right]
    \end{aligned}
\end{equation}
The scoring of the dependency $h\rightarrow m$ with label $l\in \mathcal{L}$ is calculated in a similar manner.
We use two additional MLPs and $|\mathcal{L}|$ Biaffine layers to compute all label scores.

Enhancing the first-order biaffine parser for the unlabeled tree, we leverage adjacent-sibling information as mentioned in \citet{mcdonald_online_2006} and headed-span information as described in \citet{yang_headed-span-based_2022}.  
Additional MLPs and biaffine/ triaffine layers are included to perform the scoring,
\begin{align}
&\mathbf{r}^{\mathrm{head}/\mathrm{mod}/\mathrm{sib}}_i  = \mathtt{MLP}^{\mathrm{head}/\mathrm{mod}/\mathrm{sib}}(\mathbf{h}_i)                                                  \\
       & \mathbf{s}_{h\rightarrow {s,m}}^{sib}                          =  \mathtt{TriAff}(\mathbf{r}^{\mathrm{sib}}_{s},\mathbf{r}^{\mathrm{head}}_{h},\mathbf{r}^{\mathrm{mod}}_{m})
        \\
        &\mathbf{r}^{\mathrm{left}/\mathrm{right}/\mathrm{head}}_i  = \mathtt{MLP}^{\mathrm{left}/\mathrm{right}}(\mathbf{c}_i), \mathtt{MLP}^{\mathrm{head}}(\mathbf{h}_i)                                                  \\
        &\mathbf{s}_{k,i/j}^{left/right}                          = \left[\mathbf{r}^{\mathrm{left}/\mathrm{right}}_{i/j};1\right]^{T}W\left[\mathbf{r}^{\mathrm{head}}_{k};1\right]   
\end{align}
where $s$ and $m$ are two adjacent modifiers of $h$,  $s$ populates between $h$ and $m$, $i$ and $j$ are left / right boundary of the span whose head is $k$.

Ultimately, the scoring of unlabeled tree contains the accumulation of scores of all first-order arcs, second-order adjacent-sibling arcs and headed-spans in the tree as follow,
\begin{equation}
 \begin{aligned}
    s(\boldsymbol{x}, \boldsymbol{y}) = &  \sum_{(h \rightarrow m) \in y} \mathbf{s}^{\text{arc}}_{h,m} + \sum_{(h \rightarrow \{s, m\})\in y} \mathbf{s}^{\text{sib}}_{h,s,m}\\
      &	  + \sum_{(l, r, i) \in y} (\mathbf{s}^{\text{left}}_{i, l_i} + \mathbf{s}^{\text{right}}_{i, r_i})
\end{aligned}   
\end{equation}
We parameterize the probabilities of skeletal tree $\boldsymbol{y}$ as:
\begin{equation}\label{eq:prob}
    \begin{aligned}
        P(\boldsymbol{y}\mid \boldsymbol{x})               & =\frac{\exp\left(\mathrm{s}(\boldsymbol{x},\boldsymbol{y})\right)}{Z(\boldsymbol{x})\equiv\sum_{}\exp\left(\mathrm{s}(\boldsymbol{x},\boldsymbol{y}^{\prime})\right)}                                                           \\
    \end{aligned}
\end{equation}
The term $Z(\boldsymbol{x})$ represents the partition function, which is calculable via the Inside algorithm within the TreeCRF framework.
Parameterization of the related label sequence $l$ is as follows:
\begin{equation}
        P(\boldsymbol{l}\mid\boldsymbol{x},\boldsymbol{y})  =\prod\limits_{h\xrightarrow[]{l}m \in \boldsymbol{t}} P(l\mid \boldsymbol{x},h\rightarrow m)   
\end{equation}
Each label $l$ operates independently of the tree $\boldsymbol{y}$ and other labels.
In conclusion, the probability of the labeled tree $\boldsymbol{t}$ is defined as the product of the probabilities of two components, namely, the unlabeled tree and the associated labels.
\begin{equation}
    P(\boldsymbol{t}\mid \boldsymbol{x}) = P(\boldsymbol{y}\mid\boldsymbol{x})\cdot P(\boldsymbol{l}\mid\boldsymbol{x},\boldsymbol{y})
\end{equation}

\subsection{Training Objective}

During training, the objective is to maximize the probability of tree $T_e$ and $T_{h/t}$ for each expression $e\in \mathcal{E}$.
Consequently, we formulate the loss function in the following manner:
\begin{equation}
    \mathcal{L} = - \sum_{e\in \mathcal{E}}\log P(T_e\mid\boldsymbol{x})\cdot P(T_{h/t}\mid\boldsymbol{x}) \\
\end{equation}
In this equation, the term $P(T_e\mid\boldsymbol{x})$ and $P(T_{h/t}\mid\boldsymbol{x})$ are expanded as:
\begin{equation}
    \begin{aligned}
        P(T \mid\boldsymbol{x}) & =\sum_{\boldsymbol{t}\in T} P(\boldsymbol{y}\mid \boldsymbol{x})\cdot P(\boldsymbol{l}\mid \boldsymbol{x}, \boldsymbol{y})                                                  \\
                                  & = \frac{1}{Z(\boldsymbol{x})}\sum_{\boldsymbol{t}\in T}\exp(\mathrm{s}(\boldsymbol{x},\boldsymbol{y}))\cdot P(\boldsymbol{l}\mid \boldsymbol{x},\boldsymbol{y})
    \end{aligned}
\end{equation}

\subsection{Inside Algorithm}\label{sec:inside}

The calculation of the partition function $Z(\boldsymbol{x})$ in Eq.~\eqref{eq:prob} can be resolved by the Inside algorithm of TreeCRF.
Follow \citep{zhang_semantic_2022}, we logarithm the scores and define the labeled tree score as:
\begin{equation}
    \mathrm{s}(\boldsymbol{x},\boldsymbol{t})=\mathrm{s}(\boldsymbol{x},\boldsymbol{y})+ \log P(\boldsymbol{l}\mid \boldsymbol{x},\boldsymbol{y})
\end{equation}
Consequently, the score represents the summation of the exponential scores of all legal labeled trees, as the logarithmic label probability of illegal trees is set to 0 in our conversion formulation.

To enumerate legal trees, we introduce constraints to the Inside Algorithm as proposed by \citet{eisner_bilexical_1997,li_active_2016}\footnote{We employ a modified form of \citet{yang_combining_2022} to differentiate between finished and unfinished spans}.
The constraints are categorized into two groups, each defined by its specific purpose:
(1) To prevent the arc $h\rightarrow m$ from crossing different spans, we apply a constraint to the rule (\textbf{\textsc{R-Link}}), thereby prohibiting merging with the relevant incomplete span $I_{h, m}$.
(2) To prevent the presence of multiple headwords within a single span, we restrict any word in expression spans $e$ to merge solely with the completed span $F_{h, i}$ (\textbf{\textsc{R-Comb}}). Additionally, we permit a span to be considered complete only when $i$ is positioned at the endpoint of a span (\textbf{\textsc{R-Finish}}).
We demonstrate the deduction rules via the parsing-as-deduction framework \cite{pereira_parsing_1983} in Appendix \ref{app:inside}, specifically in Figure~\ref{fig:deduction}. For additional insights into the Eisner Algorithm, Appendix \ref{app:inside} provides further details.

\section{Experiments}

\begin{table*}[!ht]
\centering
\fontsize{8.5}{9.5}\selectfont
\setlength{\tabcolsep}{4mm}
\begin{tabular}{@{}llccccc@{}}
\toprule
\multirow{2}{*}{Dataset} & 
\multirow{2}{*}{Model} & 
\multicolumn{3}{c}{Span} & 
\multicolumn{2}{c}{Sent. Graph} \\ 
\cmidrule(l){3-5}\cmidrule(l){6-7} & & 
Holder F1  & Target F1    & Exp. F1   & NSF1  & SF1 \\ 
\midrule
\multirow{7}{*}
{\textbf{NoReC}$_\textsuperscript{Fine}$} & 
RACL-BERT \citep{chen_relation-aware_2020}  & -    & 47.2   & 56.3      & -  & -  \\  
& Head-first \citep{barnes_structured_2021}            & 51.1          & 50.1          & 54.4     & 37.0             & 29.5           \\ 
& Head-final  \citep{barnes_structured_2021}            & 60.4          & \textbf{54.8} & 55.5      & 39.2           & 31.2           \\
& Frozen PERIN \citep{samuel_direct_2022}& 48.3          & 51.9         & 57.9            & 41.8           & 35.7           \\
& TGLS   \citep{shi_effective_2022}         & 60.9 & 53.2          & 61.0  & 46.4  & 37.6  \\ 
& USSA    \citep{zhai_ussa_2023}          & 66.3          & 54.3 & 61.4      & 47.7           & 39.6           \\
& Ours$^{\dagger}$           & \textbf{67.4} & 54.5& \textbf{62.7}& \textbf{49.5}& \textbf{41.5}\\
\midrule
\multirow{7}{*}{\textbf{MultiB}$_\textsuperscript{EU}$}  & 
RACL-BERT \citep{chen_relation-aware_2020}       & -             & 59.9          & 72.6           & -              & -              \\ 
& Head-first  \citep{barnes_structured_2021}            & 60.4          & 64.0            & 73.9            & 58.0             & 54.7           \\ 
& Head-final   \citep{barnes_structured_2021}           & 60.5          & 64.0            & 72.1      & 58.0             & 54.7           \\
& Frozen PERIN \citep{samuel_direct_2022}& 55.5          & 58.5          & 68.8            & 53.1           & 51.3           \\
& TGLS  \citep{shi_effective_2022}          & 62.8 & 65.6 & 75.2 &  61.1  & 58.9  \\ 
& USSA   \citep{zhai_ussa_2023}           & 63.4          & 66.9 & 75.4      & 63.5           & 60.4           \\
& Ours$^{\dagger}$           & \textbf{65.5} & \textbf{68.2}          & \textbf{75.8}& \textbf{65.7}& \textbf{62.7}\\\midrule
\multirow{7}{*}{{\textbf{MultiB}}$_\textsuperscript{CA}$}   & 
RACL-BERT \citep{chen_relation-aware_2020}             & -             & 67.5          & 70.3       & -              & -              \\ 
& Head-first  \citep{barnes_structured_2021}            & 43.0            & 72.5          & 71.1            & 62.0             & 56.8           \\ 
& Head-final \citep{barnes_structured_2021}             & 37.1          & 71.2          & 67.1            & 59.7           & 53.7           \\
& Frozen PERIN\citep{samuel_direct_2022} & 39.8          & 69.2          & 66.3            & 60.2           & 57.6           \\
& TGLS  \citep{shi_effective_2022}          & 47.4 & 73.8 & 71.8 &  64.2  & 59.8  \\ 
& USSA     \citep{zhai_ussa_2023}         & 47.5          & 74.2 & 72.2     & 67.4           & 61.0           \\
& Ours$^{\dagger}$          & \textbf{50.3} & \textbf{75.2}          & \textbf{74.7}  & \textbf{69.7}& \textbf{62.8}\\
\midrule
\multirow{7}{*}{\textbf{MPQA}} & 
RACL-BERT \citep{chen_relation-aware_2020}             & -             & 20.0            & 31.2            & -              & -              \\ 
& Head-first \citep{barnes_structured_2021}             & 43.8          & 51.0            & 48.1 &   24.5           & 17.4           \\  
& Head-final  \citep{barnes_structured_2021}            & 46.3 & 49.5          & 46.0          & 26.1           & 18.8           \\
& Frozen PERIN \citep{samuel_direct_2022}& 44.0          & 49.0          & 46.6            &30.7           & 23.1           \\
& TGLS  \citep{shi_effective_2022}          & 44.1          & 51.7 & 47.8         & 28.2  & 21.6  \\ 
& USSA  \citep{zhai_ussa_2023}            & 47.3          & 58.9 & 48.0      & 36.8          & 30.5           \\
& Ours$^{\dagger}$          & \textbf{51.2} & \textbf{60.2}          & \textbf{48.2}  & \textbf{40.1}& \textbf{32.4}\\
\midrule

\multirow{7}{*}{\textbf{DS}$_\textsuperscript{Unis}$}         & 
RACL-BERT \citep{chen_relation-aware_2020}             & -             & 44.6          & 38.2            & -              & -              \\ 
& Head-first   \citep{barnes_structured_2021}           & 28.0            & 39.9          & 40.3   &     31.0             & 25.0             \\  
& Head-final     \citep{barnes_structured_2021}         & 37.4          & 42.1          & 45.5 &  34.3           & 26.5           \\
& Frozen PERIN \citep{samuel_direct_2022}& 13.8          & 37.3          & 33.2            & 24.5           & 21.3           \\
& TGLS   \citep{shi_effective_2022}       & 43.7 & 49.0 & 42.6      & 36.1  & 31.1 \\ 
& USSA   \citep{zhai_ussa_2023}          & 44.2         & 50.2 & 46.6      & 38.0           & 33.2           \\
& Ours$^{\dagger}$           & \textbf{44.4} & \textbf{51.0}          & \textbf{48.2}  & \textbf{40.1}& \textbf{35.7}  \\\bottomrule
\end{tabular}
\caption{Main experimental results of our model and comparison with previous works. The score marked as bold means the best performance among all the methods. $\dagger$ means that the reported result is the performance of the high-order parser.
}
\label{table1}
\end{table*}
\subsection{Datasets}

Following the previous work, we conduct experiments on five benchmark datasets in four languages. 
\textbf{NoReC}$_\textsuperscript{Fine}$ \citep{ovrelid_fine-grained_2020} is a multi-domain professional reviews dataset in Norwegian. \textbf{MultiB}$_\textsuperscript{EU}$ and \textbf{MultiB}$_\textsuperscript{CA}$ \citep{barnes_multibooked_2018} are annotated hotel views in Basque and Catalan, respectively. \textbf{MPQA} \citep{wiebe_annotating_2005}contains English news and the main content of \textbf{DS}$_\textsuperscript{Unis}$ \citep{toprak_sentence_2010} is online university reviews in English as well.

\subsection{Baselines}
We compare our proposed method with six state-of-the-art baselines.
\textbf{RACL-BERT} \citep{chen_relation-aware_2020} is a relation aware framework used to resolve subtasks of ABSA coordinately.
\textbf{Head-first} and \textbf{Head-final} \citep{barnes_structured_2021} are symmetric bi-lexical dependency formulation with first-order biaffine parser.
\textbf{Frozen PERIN} \citep{samuel_direct_2022} use graph-based parser to directly extraction sentiment tuple from text.
\textbf{TGLS} \citep{shi_effective_2022} is a bi-lexical dependency parsing method applying Graph attention network to make prediction.
\textbf{USSA} \citep{zhai_ussa_2023} is the same bi-lexical method with table filling scheme for prediction.
About the details we make fair comparison, refer to Appendix \ref{app:impl} please.

\subsection{Evaluation Metrics}
Following the previous work \citep{zhai_ussa_2023}, we use \textbf{Holder F1}, \textbf{Target F1} and \textbf{Exp. F1} for the token extraction of Holders, Targets and Expressions.
For structure prediction, we use \textbf{Sentiment Graph F1 (SF1)} and \textbf{Non-polarity Sentiment Graph F1 (NSF1)} to evaluate the model, which perform exact match evaluation on the full sentiment tuple (h, t, e, p) and non-polarity tuple (h,t,e). 

\subsection{Main Results}

Table \ref{table1} shows the comparison of our method against other baselines across multiple evaluation metrics. In terms of the Span F1 metric, our method demonstrates superior performance on all datasets, including a notable 7.2\% F1 score increase in holder extraction on the MPQA dataset. Furthermore, when evaluating the Sentiment Graph metric, designed to assess both span extraction and relation prediction accuracy, our method consistently outperforms over competing approaches in both NSF1 and SF1 scores. Notably, it surpasses the USSA baseline by an average of 2.12 NSF1 and 2.08\% SF1. This enhancement is primarily attributed to our SSA as a latent tree formulation and the effective parser, which accurately models the internal structure of spans.

\subsection{Efficiency Comparison}
\begin{table}[!t]
    \newcolumntype{L}{>{\raggedright\arraybackslash}m{1.5cm}}
    \centering
    \begin{small}
        \begin{tabular}{lLr}
            \toprule \color{red!60!black}           
                & Method     & Sents/s      \\
            \midrule
        Head-first \citep{barnes_structured_2021} & \textsc{1O}  & 170       \\
        TGLS \citep{shi_effective_2022}   & \textsc{TF}   & 79  
        \\
        USSA \citep{zhai_ussa_2023} & \textsc{TF}   & 50           \\
         \multirow{2}{*}{Ours}   & \textsc{1O}      & \textbf{212} \\
                                 & \textsc{2O \& span} & 174          \\                            
            \bottomrule
        \end{tabular}
    \end{small}
    \caption{
        Speed comparison on MPQA Test dataset. The dataset contains 1681 sentences with average 24 tokens. TF:Table-filling method, 1O:first-order parser, 2O:second-order parser.
    }
    \label{table:speed}
\end{table}

Table \ref{table:speed} presents a comparative analysis of different models based on their processing speeds. The speed metrics for previous works were derived by rerunning their publicly available code. Our models, the first-order scoring only method and the comprehensive scoring method, demonstrate superior performance, processing approximately 212 and 174 sentences per second, respectively. This rate significantly surpasses that of prior models. TGLS \citep{shi_effective_2022} and USSA \citep{zhai_ussa_2023} employ deeper networks and engage in computationally intensive tasks, 
which contributes to their slower processing speeds.

\section{Discussion}
\label{sec:discussion}

\subsection{Can SSA as latent tree formulation handle overlap and discontinuous cases?}

SSA as a latent tree formulation effectively addresses not only latent tree modeling but also the management of overlap and discontinuous scenarios, as highlighted by \citet{zhai_ussa_2023}. We offer a succinct proof as follows:
(1) \textbf{Overlap Case}: Our approach handles overlaps efficiently by treating each expression separately. This allows different expressions to share the same target/holder without conflict in our framework.
(2) \textbf{Discontinuous Case}: This more complex scenario, predominantly found in expressions\footnote{Discontinuity occurs in approximately 2\% of targets and holders and 4\%-9\% of expressions across five benchmarks, excluding the MPQA dataset, which lacks discontinuous cases. See \citet{zhai_ussa_2023} for detailed statistics.}, is also addressed in our scheme. We introduce a new label, ``\textbf{exp-incomplete}'', during the expression extraction subtask. This label is based on the observation that discontinuous cases often include modifiers like adverbs of degree, affirmation, probability, or negation, which can be identified independently. For instance, in ``It is \textbf{by no means} a \textbf{diploma mill}'', the discontinuous expression (``by no means'' and ``diploma mill'') is handled by recognizing ``by no means'' as an incomplete negation, which, when attached to a complete expression like ``diploma mill'' (negative), alters the overall sentiment polarity to positive in our framework.\footnote{It's important to note that incomplete expression spans do \textbf{not} correspond to specific target/holder pairs; they are solely linked to complete expressions.}

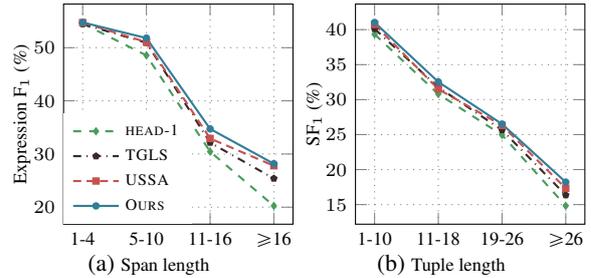
\begin{figure}[tb!]
    \centering
    \begin{small}
        \captionsetup[subfigure]{aboveskip=-1pt,belowskip=-1pt}
        \begin{subfigure}[t]{0.49\columnwidth}
            \centering
            \begin{tikzpicture}
                \begin{axis}[
                        legend style={
                                draw=none,
                                fill opacity=2,
                                text=black,
                                at={(0,0)},
                                anchor=south west,
                                font=\tiny
                            },
                        symbolic x coords={1-4,5-10,11-16,$\geq$16},
                        xtick=data,
                        font=\scriptsize,
                        ylabel={Expression F$_1$ (\%)},
                        ylabel style = {yshift=-5pt},
                        legend cell align={left},
                        height=4.5cm,
                        width=4.6cm,
                        grid=major,
                        grid style={dotted, color=gray}
                    ]
                    \addplot[mark=diamond*, mark options={solid,mark size=1pt}, dashed,thick,forestgreen] coordinates {
                            (1-4,     54.54)
                            (5-10,     48.55)
                            (11-16,     30.45)
                            ($\geq$16, 20.25)
                        };
                    \addlegendentry{\textsc{head-1}}
                    \addplot[mark=pentagon*, mark options={solid,mark size=1pt}, dashdotted,thick,burntorange] coordinates {
                            (1-4,     54.45)
                            (5-10,     50.93)
                            (11-16,     32.14)
                            ($\geq$16, 25.40)
                        };
                    \addlegendentry{\textsc{TGLS}}
                    \addplot[mark=square*, mark options={solid,mark size=1pt}, densely dashed,thick,brickred!80] coordinates {
                            (1-4,     54.73)
                            (5-10,     51.04)
                            (11-16,     32.92)
                            ($\geq$16, 27.79)
                        };
                    \addlegendentry{\textsc{USSA}}
                    \addplot[mark=*, mark options={solid,mark size=1pt}, thick,midnightblue!80] coordinates {
                            (1-4,     54.75)
                            (5-10,     51.80)
                            (11-16,     34.72)
                            ($\geq$16, 28.17)
                        };
                    \addlegendentry{\textsc{Ours}}
                \end{axis}
            \end{tikzpicture}
            \caption{\scriptsize Span length}
            \label{fig:f1-arg-len}
        \end{subfigure}
        \hfill
        \begin{subfigure}[t]{0.49\columnwidth}
            \centering
            \begin{tikzpicture}
                \begin{axis}[
                        font=\scriptsize,
                        symbolic x coords={1-10,11-18,19-26,$\geq$26},
                        xtick=data,
                        ytick = {15, 20, ..., 30,35,40},
                        ylabel={SF$_1$ (\%)},
                        ylabel style = {yshift=-5pt},
                        legend cell align={left},
                        height=4.5cm,
                        width=4.6cm,
                        grid=major,
                        grid style={dotted, color=gray}
                    ]
                    \addplot[mark=diamond*, mark options={solid,mark size=1pt}, dashed,thick,forestgreen] coordinates {
                            (1-10,     39.33)
                            (11-18,     30.80)
                            (19-26,     24.95)
                            ($\geq$26, 14.82)
                        };
                    \addplot[mark=pentagon*, mark options={solid,mark size=1pt}, dashdotted,thick,burntorange] coordinates {
                            (1-10,     40.11)
                            (11-18,     31.76)
                            (19-26,     25.66)
                            ($\geq$26, 16.36)
                        };
                    \addplot[mark=square*, mark options={solid,mark size=1pt}, densely dashed,thick,brickred!80] coordinates {
                            (1-10,     40.74)
                            (11-18,     31.50)
                            (19-26,     26.31)
                            ($\geq$26, 17.33)
                        };
                    \addplot[mark=*, mark options={solid,mark size=1pt}, thick,midnightblue!80] coordinates {
                            (1-10,     41.03)
                            (11-18,     32.53)
                            (19-26,     26.51)
                            ($\geq$26, 18.21)
                        };
                \end{axis}
            \end{tikzpicture}
            \caption{\scriptsize Tuple length}
            \label{fig:f1-dist}
        \end{subfigure}
    \end{small}
    \caption{
        Analysis on long span identification in NoReC Test Dataset. The dataset contains 8448 expressions with up to 30 token and average length of 4.9 tokens. The Head-1 represents head-first method. (a) Expression F$_1$ scores breakdown by span length  (b) SF$_1$ scores breakdown by tuple length.
    }
    \label{fig:f1-breakdown}
\end{figure}

\begin{table}[t]
\centering
\setlength{\tabcolsep}{0.5mm}
\begin{tabular}{@{}lcccc@{}}
\toprule
 & Holder F1 & Target F1 & Exp. F1 & SF1  \\ 
 \midrule
\textbf{Ours}    & \textbf{51.2}  & \textbf{60.2} &\textbf{48.2}  &\textbf{32.4} \\
\hdashline
\textcolor{white}{pad}First     & 48.5     & 57.4       & 48.1 &  27.8\\
\textcolor{white}{pad}Last      & 48.6     & 57.2      & 48.0     & 27.5 \\
\textcolor{white}{pad}Flat     & 45.4      &  53.3    & 47.6     & 24.1 \\
 \bottomrule
\end{tabular}
\caption{Experimental results of our first-order method and three variants methods \textbf{Flat}, \textbf{First} and \textbf{Final} with various degrees of excluding internal structures in MPQA Test dataset.}
\label{tab:latent}
\end{table}

\subsection{Does the latent tree structure benefit for long spans/tuples?}

We address the question through results from two key experiments:
(1) Experiment on \textbf{NoReC}$_\textsuperscript{Fine}$ (Figure~\ref{fig:f1-breakdown}) investigates the model's performance in longer spans/tuples. The Expression F1 and Tuple F1 metrics indicate our method's superiority over baselines, with an increasing performance gap correlating with longer span/tuple lengths.
(2) To assess the importance of modeling spans as latent trees, we examine three variants of our first-order method:
1) \textsc{First}: Utilizes the first word as the span headword.
2) \textsc{Last}: Employs the last word as the span headword.
3) \textsc{Flat}: Similar to the head-first method by \citet{barnes_structured_2021}, attaches span words directly to the first word.
Results in Table~\ref{tab:latent} reveal that both \textsc{First} and \textsc{Last} perform comparably but are consistently outperformed by our first-order method. \textsc{Flat} records a significant drop in performance (24.1 SF1 on \textbf{MPQA}) compared to our method (32.4 SF1), indicating the substantial contribution of latent span representations to our model's effectiveness.

\section{Conclusion}

In this study, we approach Structured Sentiment Analysis (SSA) through latent dependency graph parsing, conceptualizing flat sentiment spans as latent subtrees. We introduce an innovative parsing methodology grounded in TreeCRF, designed to effectively integrate span structures. Our experimental findings demonstrate that this method surpasses all previous approaches across five benchmark datasets. Comprehensive analyses validate the efficacy and consistency of our method in enhancing SSA.

\section*{Acknowledgments}
This work is supported by the National Natural Science Foundation of China (No. 62176187).

\section*{Limitations}
We propose a two-stage parsing method to model the converted latent tree derived from the original SSA structure.
The experiments demonstrate that our method outperforms the previous methods on benchmark datasets and proves its temporal efficiency.
However, inevitable error propagation occurs in our decoding stage due to the sequential nature of the two-stage decoding process.
Another challenge is that considering each expression span separately to construct their corresponding trees incurs increased space requirements for storing intermediate results.

\section*{Ethics Statement}
Our work on the Structured Sentiment Analysis (SSA) as Latent Graph Parsing Scheme adheres to ethical guidelines emphasizing transparency, fairness, and responsible AI development. 
We recognize the ethical implications of this work and have conducted our research with a commitment to minimizing biases, ensuring data privacy, and promoting the explainability of AI decisions. 
Our evaluations utilized publicly available or ethically sourced datasets, and we have made efforts to address and mitigate potential biases within these datasets to ensure fairness and objectivity in our findings.

The broader impact of Latent Graph Parsing Scheme, aimed at improving sentiment analysis for a more complete conceptualization, has the potential to contribute positively to various fields, including negation resolution, uncertainty/hedge detection, and event extraction. 
By introducing a graph parsing method, we foster more faithful, flexible, and explainable sentence-level sentiment analysis capabilities.
We encourage the responsible use of our findings and technologies, and we commit to ongoing evaluation of our work’s societal and ethical implications.

\bibliography{main}
\bibliographystyle{acl_natbib}

\newpage

\appendix
\section{Implement Details}
\label{app:impl}
For fair comparison, we obtain the contextual token representations from the pre-trained BERT-multilingual-base and word2vec skip-gram embeddings used by previous work \citep{barnes_structured_2021}. Furthermore, we use 4-layer BiLSTMs to encoder the sentence,with an output size of 768 and the dropout rate is set to 0.3. We train our model for 60 epochs and save the model parameters based on the highest SF1 score on the development set. We conduct training and testing on a NVIDIA A100 Server. 
The reported results are the averages from five runs with different random seeds. Our proposed model statistically outperforms the baselines at p < 0.05.

\section{Inside algorithm}
\label{app:inside}

\begin{figure}[ht!]
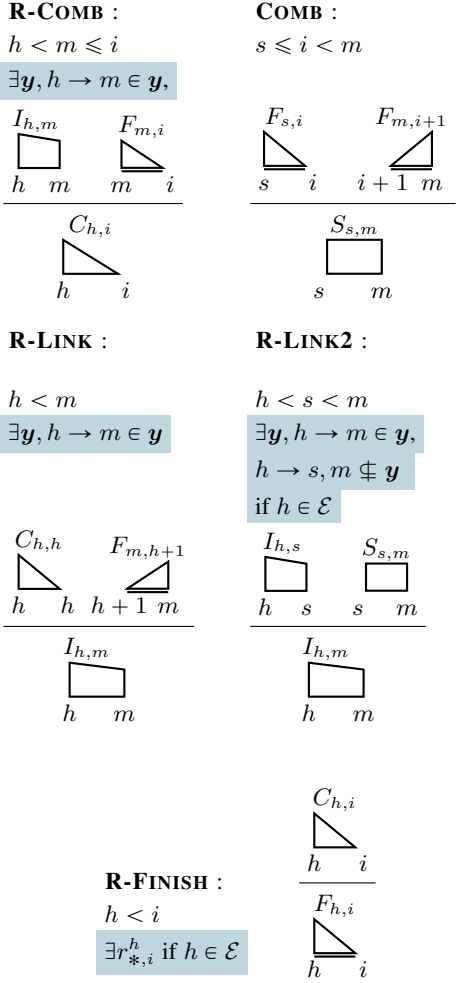

    \centering
    
    \begin{subfigure}{0.9\columnwidth}
        
    \renewcommand{\arraystretch}{1}
    \centering
    \small
    \begin{tabular}{ll}
        \textbf{\colorbox{white}{\textsc{R-Comb}}}:\hspace{4em} & 
        \textbf{\colorbox{white}{\textsc{Comb}}}:  
        \\
        \begin{tabular}[t]{@{}l@{}} \colorbox{white}{$h < m \leq i$}   \\ \colorbox{midnightblue!25}{$\exists \boldsymbol{y},h\rightarrow m \in \boldsymbol{y}$,}\\ \end{tabular} 
        &
        \begin{tabular}[t]{@{}l@{}} \colorbox{white}{$s \leq i <m$} \\  \end{tabular}
        \\[2pt]
        $\inferrule{
                \tikz[baseline=0pt]{\trapezoidd[0.6][0.5][0.1]{h}{m}{I_{h,m}}} \quad
                \tikz[baseline=2.5pt]{\rightclosetrianglee[0.6][0.4]{m}{i}{F_{m,i}}}
            } { \tikz[baseline=0pt]{\righttrianglee[0.8][0.5]{h}{i}{C_{h,i}}} }$  &
        $\inferrule{
                \tikz[baseline=0pt]{\rightclosetrianglee[0.6][0.5]{s}{i}{F_{s,i}}} \quad
                \tikz[baseline=0pt]{\leftclosetrianglee[0.6][0.5]{i+1}{m}{F_{m,i+1}}}
            } { \tikz[baseline=0pt]{\square[0.8][0.5]{s}{m}{S_{s,m}}} }$

        \\[40pt]
        \textbf{\colorbox{white}{\textsc{R-Link}}}:    & 
        \textbf{\colorbox{white}{\textsc{R-Link2}}}: \\
        \\
        \begin{tabular}[t]{@{}l@{}}\colorbox{white}{$h < m$} \\ \colorbox{midnightblue!25}{$\exists \boldsymbol{y}, h\rightarrow m \in \boldsymbol{y}$} \end{tabular}     &
        \begin{tabular}[t]{@{}l@{}}
        \colorbox{white}{$h< s < m$} \\
        \colorbox{midnightblue!25}{$\exists \boldsymbol{y},h\rightarrow m \in \boldsymbol{y}$,}\\
        \colorbox{midnightblue!25}{$h\rightarrow s,m \nsubseteq  \boldsymbol{y}\;$}\\
        \colorbox{midnightblue!25}{$\textrm{if}\; h \in \mathcal{E}$}
        \end{tabular}  \\[2pt]
        $\inferrule{
                \tikz[baseline=0pt]{\righttrianglee[0.6][0.5]{h}{h}{C_{h,h}}}
                \tikz[baseline=2.5pt]{\leftclosetrianglee[0.6][0.4]{h+1}{m}{F_{m,h+1}}}
            } {
                \tikz[baseline=0pt]{\trapezoidd[0.8][0.5][0.1]{h}{m}{I_{h,m}}}
            }$   &
        $\inferrule{
                \tikz[baseline=0pt]{\trapezoidd[0.6][0.5][0.1]{h}{s}{I_{h,s}}} \quad
                \tikz[baseline=2.5pt]{\square[0.6][0.4]{s}{m}{S_{s,m}}}
            } {
                \tikz[baseline=0pt]{\trapezoidd[0.8][0.5][0.1]{h}{m}{I_{h,m}}}
            }
        $
        \\[40pt]

        \\[2pt]
    \end{tabular}
    \begin{tabular}{cc}
         
    \begin{tabular}[t]{l}  
    \textbf{\colorbox{white}{\textsc{R-Finish}}}: \\
    \colorbox{white}{$h <i$} \\ 
    \colorbox{midnightblue!25}{$\exists r^h_{*,i}\;\textrm{if}\; h \in \mathcal{E}$} \end{tabular} 
    
    & 
        $\inferrule{
                \tikz[baseline=0pt]{\righttrianglee[0.6][0.5]{h}{i}{C_{h, i}}} 
            } {
                \tikz[baseline=0pt]{\rightclosetrianglee[0.6][0.5]{h}{i}{F_{h, i}}}
            }
        $
        \\[2pt]
    \end{tabular}
    
    \end{subfigure}
    
    \caption{
    Deduction rules for our constrained Inside algorithm based on Eisner algorithm \citep{eisner_bilexical_1997}, we extend it to second-order and introduce span scoring \citep{yang_combining_2022} for unify the deduction framework to expression identification and expression-target/holder relation prediction.
    (\textbf{\textsc{R-Comb}} and \textbf{\textsc{R-Link}}), combination with span scoring (\textbf{\textsc{R-Finish}}) and its second-order extension (\textbf{\textsc{Comb}} and \textbf{\textsc{R-Link2}}).
    Our modified rule constraints are highlighted in \emph{blue} color.
    $r^h_{*,i}$ denotes a span that takes $h$ as the headword and ends with $i$, $h \in \mathcal{E}$ denotes h is a word within expression span.
    We show only R-rules, omitting the symmetric L-rules as well as initial conditions for brevity.
    }
    \label{fig:deduction}
\end{figure}

We give the pseudocode of the modified Inside algorithm \cite{yang_combining_2022} in Alg.~\ref{alg:inside-2o} as well as the illustration of deduction rules in Figure~\ref{fig:deduction} to help understanding the algorithm.
We highlight the constraint rules in Figure~\ref{fig:deduction},
it is the only difference between our proposed one and the vanilla inside algorithm.
\begin{algorithm}[tb!]
    \begin{algorithmic}[1]
        \newlength{\commentindent}
        \setlength{\commentindent}{.24\textwidth}
        \STATE \textbf{Define:} $I, S, C, F \in \mathbf{R}^{n \times n \times b}$
        \STATE $\;$ 
        \STATE \textbf{Initialize:} $C_{i, i} = \mathbf{0}, 0 \le i \le n$\label{alg:init}
        \FOR {$w = 1$ \TO $n$}
        \STATE \emph{Parallelization on} $0 \le i$; $j=i+w \le n$
        
        \STATE $F_{i, j} \leftarrow C_{i, j} +  \mathrm{s}_{i, j}^{right}$\label{alg:finished-r}
        \STATE $F_{j, i} \leftarrow C_{j, i} +  \mathrm{s}_{j, i}^{left}$\label{alg:finished-l}
        \STATE $\begin{aligned}
                I_{i, j} & \leftarrow \log(\exp(C_{i, i} + F_{j, i+1})                         \\
                         & + \sum_{i < r < j} \exp(I_{i, r} + S_{r, j} + \mathrm{s}_{i, r, j}^{sib})) \\
                         & + \mathrm{s}_{i, j}^{arc}
            \end{aligned} $\label{alg:incomplete-r}
        \STATE $\begin{aligned}
                I_{j, i} & \leftarrow \log(\exp(C_{j, j} + F_{i, j-1})                         \\
                         & + \sum_{i < r < j} \exp(I_{j, r} + S_{r, i} + \mathrm{s}_{i, r, j}^{sib})) \\
                         & + \mathrm{s}_{j, i}^{arc}
            \end{aligned} $\label{alg:incomplete-l}

        \STATE $S_{i, j} \leftarrow \log\sum_{i \le r < j} \exp(F_{i, r} +  F_{j, r+1})$\label{alg:sib}
        \STATE $C_{i, j} \leftarrow \log\sum_{i < r \le j} \exp(I_{i, r} +  F_{r, j})$\label{alg:complete-r}
        \STATE $C_{j, i} \leftarrow \log\sum_{i \le r < j} \exp(I_{j, r} +  F_{r, i})$\label{alg:complete-l}

        \ENDFOR
        \RETURN $F_{0, n}$
    \end{algorithmic}
    \caption{The Second-order Inside Algorithm.}
    \label{alg:inside-2o}
\end{algorithm}

In Line~\ref{alg:init}, the term $C_{i,i}$ denotes the axiom items $\tikz[baseline=-10pt]{\righttrianglee[0.6][0.3]{i}{i}{}}$ assigned an initial score of $\mathbf{0}$. Line~\ref{alg:finished-r} is associated with the complete span $C_{i,j}$, assuming all child spans with the right boundary of $j$ are absorbed. This is achieved by adding the scores of headed-split right spans to derive the finished span $F_{i,j}$ ($\tikz[baseline=-10pt]{\rightclosetrianglee[0.6][0.3]{i}{j}{}}$) as  \textbf{\textsc{R-Finish}} operation. Line~\ref{alg:incomplete-r} aligns with two merging processes depicted in Figure~\ref{fig:deduction}. The incomplete span $I_{i,j}$ ($\tikz[baseline=-10pt]{\trapezoidd[0.6][0.3][0.1]{i}{j}{}}$) emerges from aggregating either complete span $C_{i,i}$ with Finished span $F_{j,i+1}$ (\textbf{\textsc{R-Link}}) or combining incomplete span $I_{i,r}$ with sibling span $S_{j,r}$ (\textbf{\textsc{R-Link2}}). In Line\ref{alg:sib}, the sibling span $S_{i,j}$ ($\tikz[baseline=-10pt]{\square[0.6][0.3]{i}{j}{}}$) results from summing finished spans $F_{i,r}$ and $F_{j,r+1}$ (\textbf{\textsc{Comb}}). Line~\ref{alg:complete-r} details a merging operation for pairs of incomplete span $I_{i,r}$ and finished span $F_{r,j}$ to form a complete span $C_{i,j}$ ($\tikz[baseline=-10pt]{\righttrianglee[0.6][0.3]{i}{j}{}}$) as \textbf{\textsc{R-Comb}} rule. Lines~\ref{alg:incomplete-l} and \ref{alg:complete-l} describe symmetric L-rules, which are not included in Figure~\ref{fig:deduction}.

\section{Comparing to Large Language Models}
\label{app:llm}

\begin{table}[!h]
\centering
\fontsize{8.5}{9.5}\selectfont
\setlength{\tabcolsep}{0.45mm}
\begin{tabular}{@{}lccccc@{}}
\toprule
\multicolumn{1}{l}{} & \textbf{NoReC}$_\textsuperscript{Fine}$ & \textbf{MultiB}$_\textsuperscript{EU}$   & \textbf{MultiB}$_\textsuperscript{CA}$   & \textbf{MPQA }                  & \textbf{DS}$_\textsuperscript{Unis}$   \\ \midrule
Ours & \textbf{41.5} & \textbf{62.7} &\textbf{62.8} &\textbf{32.4} &\textbf{35.7}\\
GPT 3.5           & 27.3  & 29.5 & 32.3 & 20.6 & 22.5 \\ \cdashline{1-6}
GPT 4 & 32.3& 34.2&35.2 & 24.5& 25.6\\
\textcolor{white}{pad}+1 shot          & 35.5      & 37.5     & 38.5     &  \underline{25.2}                      &  \underline{27.6}    \\ 
\textcolor{white}{pad}+5 shot               & \underline{35.8}  & \underline{38.1} & \underline{39.8} & 24.6                   & 23.4 \\ \bottomrule
\end{tabular}
\caption{Experimental results of the SF1 score utilize the ChatGPT and GPT4 to generate the sentiment tuple of five benchmarks. We underline the best results of GPT performance, which still lags behind our method (the highlight ones) largely.}
\label{tab:gpt}
\end{table}
In Table \ref{tab:gpt}, we present a comparative analysis of our method against ChatGPT-3.5-Turbo and GPT4. The results indicate that on the English dataset, GPT4 surpasses the baseline performance in a zero-shot setting and shows further improvement with in-context learning (1-shot or more). However, GPT4 does not reach the performance levels of more robust baselines such as TGLS and USSA, which are specifically trained for the SSA task.

\end{document}